\documentclass[journal]{IEEEtran}

\usepackage{graphics} 
\usepackage{epsfig} 
\usepackage{cite} 
\usepackage{multirow}
\usepackage{fancyhdr}
\usepackage{hyperref}
\ifCLASSINFOpdf
\else
\fi
\usepackage{amsmath}
\usepackage{amsthm,amsmath,amssymb}
\usepackage{mathrsfs}

\DeclareMathOperator*{\argmax}{arg\,max}
\DeclareMathOperator*{\CE}{CE}

\usepackage{algorithm}
\usepackage{algpseudocode}
\usepackage{algorithmicx}

\hypersetup{urlcolor=magenta,colorlinks=true,linkcolor=black}

\begin{document}
\title{Conservative-Progressive Collaborative Learning for Semi-supervised Semantic Segmentation}

\author{Siqi Fan, Fenghua Zhu, Zunlei Feng, Yisheng Lv, Mingli Song, Fei-Yue Wang

\thanks{Siqi Fan is with the State Key Laboratory for Management and Control of Complex Systems, Institute of Automation, Chinese Academy of Sciences, Beijing, China, and also with the Institute for AI Industry Research (AIR), Tsinghua 
University, Beijing, China.
        {\tt\small (fansiqi2019@ia.ac.cn)}}
\thanks{Fenghua Zhu, Yisheng Lv, and Fei-Yue Wang are with the State Key Laboratory for Management and Control of Complex Systems, 
Institute of Automation, Chinese Academy of Sciences, Beijing, China.
        {\tt\small (fenghua.zhu@ia.ac.cn; yisheng.lv@ia.ac.cn; feiyue.wang@ia.ac.cn)}}

\thanks{Zunlei Feng and Mingli Song are with the College of Computer Science and Technology, Zhejiang University, Hangzhou, China.
{\tt\small (zunleifeng@zju.edu.cn; brooksong@zju.edu.cn)}}
\thanks{Corresponding author: 
Fenghua Zhu \& Zunlei Feng. Code will be made publicly available upon publication.}

}



\maketitle

\begin{abstract}

  Pseudo supervision is regarded as the core idea in semi-supervised learning for semantic segmentation, and there is always a tradeoff between utilizing 
  only the high-quality pseudo labels and leveraging all the pseudo labels. Addressing that, we propose a novel learning approach, called 
  Conservative-Progressive Collaborative Learning (CPCL), among which two predictive networks are trained in parallel, and the pseudo supervision is 
  implemented based on both the agreement and disagreement of the two predictions. One network seeks common ground via intersection supervision and is 
  supervised by the high-quality labels to ensure a more reliable supervision, while the other network reserves differences via union supervision and is 
  supervised by all the pseudo labels to keep exploring with curiosity. Thus, the collaboration of conservative evolution and progressive exploration can 
  be achieved. To reduce the influences of the suspicious pseudo labels, the loss is dynamic re-weighted according to the prediction confidence. Extensive 
  experiments demonstrate that CPCL achieves state-of-the-art performance for semi-supervised semantic segmentation.

\end{abstract}


\IEEEpeerreviewmaketitle

\section{Introduction}

\IEEEPARstart{I}{mage} semantic segmentation is a core computer vision task, and plays an important role for the environment perceptions of many 
intelligent systems \cite{zhu2019parallel}. With the wide adoption of deep supervision learning, current state-of-the-art methods have performed satisfactory predictions 
given sufficient pixel-level labeled data. However, the performance would drop significantly when the labeled data is limited. It is well-known 
that the acquisition of the large amount of pixel-level labeled data for segmentation is costly and time consuming. For example, the annotation 
process costs more than 1.5h for a human annotator to label a high-resolution image of urban street scenes \cite{Cityscapes}, which is 15 times 
and 60 times larger than that of region-level and image-level labels \cite{COCO}. Therefore, this paper study semi-supervised learning (SSL) for 
semantic segmentation which leverages the large amount of unlabeled data to alleviate the labeled data dependency.

\begin{figure}[t]
  \centering
  \includegraphics[scale=0.5]{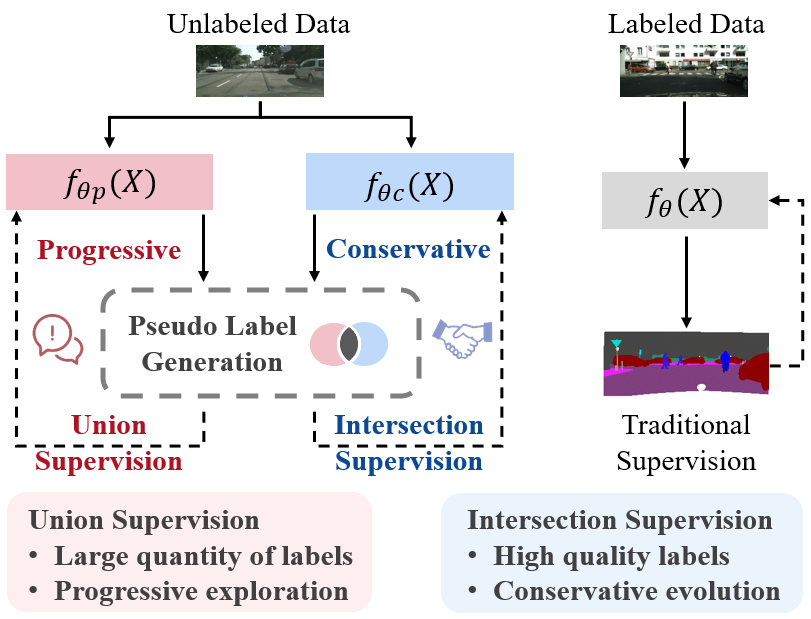}
  \caption{Diagram of the Conservative-Progressive Collaborative Learning (CPCL). There are two paralleled branches, one is conservative, and the other is 
  progressive. The conservative branch is under the intersection supervision by using high-quality pseudo labels, which achieves the conservative evolution 
  based on the agreement. The progressive branch is supervised by the union supervision utilizing the large quantity of labels, which achieves the progressive 
  exploration for the disagreement. The collaborative learning of the conservative and progressive contributes to the effective exploitation and exploration 
  of the unlabeled data. Code is available at \href{https://github.com/leofansq/CPCL}{GitHub}.}
  \label{fig:AD}
\end{figure}

\begin{figure*}[h]
  \centering
  \includegraphics[scale=0.65]{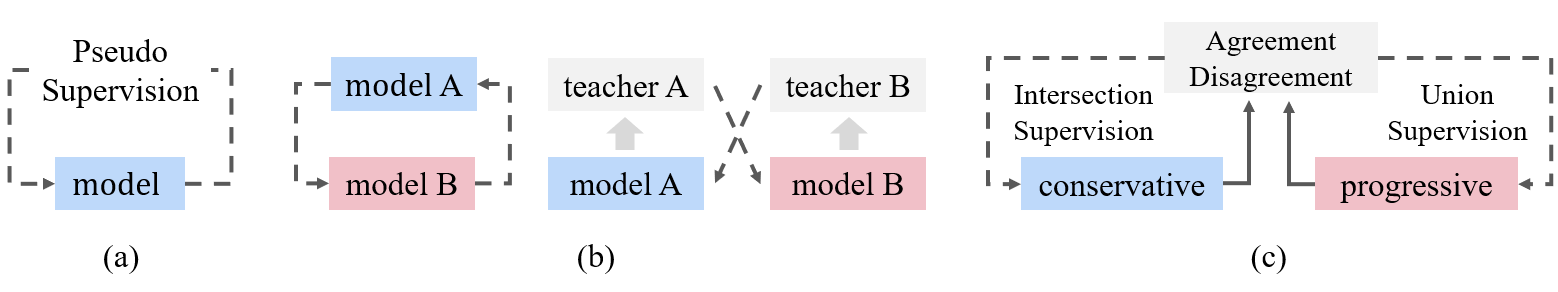}
  \caption{Comparison of (a) single-branch self-training, (b) typical double-branch mutual learning and (c) the proposed conservative-progressive collaborative learning.}
  \label{fig:comparison}
\end{figure*}

Though SSL is still in its initial stage, some pioneering works have been accomplished. Entropy minimization and consistency 
regularization are two typical approaches which are proposed in recent years. Entropy minimization methods can generate the pseudo labels on unlabeled images 
using the model trained on the labeled images, and retrain the model with the expand training data \cite{SSLEM,ALS4}. Consistency regularization methods aim 
to enforce an invariance of the predictions with various perturbations \cite{MT,VAT,9301261}, including the perturbation on input images \cite{2019Semi}, 
features \cite{CCT} and network \cite{DualStudent}. During the process of pseudo supervision, both the two lines of approaches are plagued with confirmation 
bias \cite{PLCB}, which are caused by the errors in pseudo labels. Addressing that, prediction confidence is adopted by a few recent works as indicator to 
select reliable pseudo labels \cite{ALS4,2021In}. Though setting the threshold of the confidence can avoid performance degradation, a large portion of 
unlabeled data is wasted. For example, only 27\%-36\% pixels can be used in the experiments of Cityscapes \cite{ALS4}, which is far from the goal of SSL. 
There is always a tradeoff between quality and quantity. Besides prediction confidence, the agreement of multiple predictions can also be used as an 
indicator, which is the basic idea of some mutual learning methods \cite{ke2019dual,qiao2018deep, 2020DMT}. Teacher-Student and Student-Student are two 
typical mutual learning manners, and they are both faced with challenges. The inequality in Teacher-Student may cause misleading and limitation, while the 
two networks trained in Student-Student manner may get into model coupling problem that reduces the mutual learning to self-training and hinders the further 
improvement \cite{RML}.

Considering these limitations, we propose the \textbf{Conservative -Progressive Collaborative Learning (CPCL)}, which trains two paralleled networks via the 
intersection supervision using only high-quality pseudo labels and union supervision using large-quantity pseudo labels, respectively, and achieves the 
collaboration of conservative evolution and progressive exploration, as shown in Fig.~\ref{fig:AD}. The two networks share the same structure, but are 
initialized differently. The pseudo labels are generated according to the predictions of the two networks on the unlabeled images. Specifically, one network 
is conservative and supervised by the more reliable `intersection' parts, while the other network is progressive and is under the pseudo supervision of the 
`union' parts to explore the disagreement predictions. Therefore, the two networks are trained under the heterogeneous knowledge, and the model coupling 
problem can be reduced to a certain extent. In addition, the prediction confidence is used for loss re-weighting to deal with the ineluctable noisy pseudo 
labels.

The main contributions of our work are listed as follows.

\begin{itemize}
  \item We propose the Conservative-Progressive Collaborative Learning (CPCL), which seeks common ground via the intersection supervision while reserving differences 
  through the union supervision, and achieves the collaboration of conservative evolution and progressive exploration. The intersection supervision utilizes 
  the high-quality pseudo labels, and the union supervision leverages the large quantity of pseudo labels. The two paralleled networks are trained using heterogeneous 
  knowledge to reduce the model coupling problem.

  \item The class-wise disagreement indicator and the confidence -based dynamic loss are proposed. The indicator is proposed and calculated to determine the pseudo 
  labels of the disagreement parts. The confidence-based dynamic loss utilizes prediction confidence for loss re-weighting, instead of thresholding.
  
  \item Extensive experimental studies are conducted with various settings on two widely used semantic segmentation datasets, Cityscapes and PASCAL VOC 2012. 
  CPCL shows state-of-the-art performance for semi-supervised segmentation on two benchmarks, and the effectiveness of our approach is verified.
\end{itemize}

\section{Related Work}

In this paper, we focus on how to leverage the available unlabeled images to derive additional training signals. From the perspective of architecture, the comparison 
of the single-branch self-training, two typical double-branch mutual learning approaches and the proposed conservative-progressive collaborative learning is shown in 
Fig.~\ref{fig:comparison}. Different from others, the two branches are equally trained using heterogeneous knowledge to reduce the misleading and model coupling problem. 
From the viewpoint of the prior knowledge, SSL methods are mainly based on three types knowledge, including entropy minimization, consistency regularization and 
agreement-based knowledge.

\subsection{Entropy Minimization}

Entropy minimization encourages the model to make low-entropy predictions on unlabeled data. The predictive entropy for a model is defined as
\begin{equation}
  H = \sum_{c=1}^C p_c \, \log \, p_c ,
\end{equation}
where $C$ is the number of classes and $p_c$ is the probability for class $c$. The entropy would decrease when the model makes confident predictions on unlabeled data. 
Self-training based on pseudo labeling is a typical technique for entropy minimization, which utilizes the pseudo labels for fine-tuning the segmentation model. The pseudo 
supervision of the online self-training methods is propagated backward immediately \cite{Pseudo-label}, while the relabeling-retrain process of the offline self-training 
methods can be iterated several times \cite{zou2018unsupervised, Naive-Student, rethinking, 2020DMT, ALS4, ibrahim2020semi, mendel2020semi}. The quality of the pseudo labels 
is the key of entropy minimization, and most of methods select the predictions with higher confidence values as the pseudo labels. Earlier works 
regard the predicted probability as the prediction confidence \cite{li2017triple, luo2018smooth, springenberg2015unsupervised}, and some of them 
set hard threshold for selection \cite{Pseudo-label,zou2018unsupervised}. The pseudo label $l$ is defined as 
\begin{eqnarray}
  l = 
  \begin{cases}
    \overline{c} & f(\overline{c}|x)>T\\
    ignored & otherwise
  \end{cases} ,
\end{eqnarray}
where $\overline{c} = \argmax_{c} f(c|x)$, and $T$ is the threshold. The GAN-based methods utilize the discriminator as the selector to 
approximate the confidence, and select high-confident predictions by distinguishing the ground-truth and predictions \cite{ALS4, souly2017semi, 8935407}. Other works regard 
averaged predictions as pseudo labels based on the assumption that the averaged predictions are more confident. FastSWA \cite{athiwaratkun2018there} averages models between 
epochs, while Mean-Teacher \cite{MT} turns to exponential moving average. Temporal Model \cite{laine2016temporal} leverages the predictions over epochs. These methods assume 
that the labeled data is enough to let the trained model produce the pseudo labels, which is invalid in low-data regime, and the pseudo labels with noise might result in 
self-error accumulation via training. 

\subsection{Consistency Regularization}
Consistency regularization enforces the model remain unchanged when various perturbations occur. Input perturbations \cite{kim2020structured} generate the inconsistency by 
altering the inputs, which is similar to data augmentation. Data interpolation is adopted by MixMatch \cite{MixMatch} and ReMixMatch \cite{berthelot2019remixmatch}. 
CutMix \cite{2019Semi} extends Cutout \cite{devries2017improved} and Mixup \cite{zhang2017mixup} by replacing the image regions with patches from another image, and is used 
in several methods \cite{RML, CPS}. VAT \cite{VAT} generates adversarial perturbations that alter the predictions most. FixMatch \cite{sohn2020fixmatch} and 
PseudoSeg \cite{pseudoseg} encourage the consistency of the predictions under strongly-augmented and the weakly-augmented. Feature perturbations are adopted by 
CCT \cite{CCT}, which alters the encoder's outputs and enforces the consistency between the predictions of multiple auxiliary decoders. Other works either leverage the Dropout 
to perturb the predictions \cite{park2018adversarial, laine2016temporal}, or perform network perturbations using two same and independently-initialized network 
branches \cite{GCT, CPS}. The performance of these methods heavily depends on the perturbations they used.

\subsection{Mutual Learning Based on Agreement and Disagreement}
Mutual learning trains two networks in parallel, which enables the utilization of multiple predictions to generate the pseudo supervision. Dual Student \cite{ke2019dual} 
trains two networks online, and one teaches another if it has more certain prediction. The two networks in Deep Co-training \cite{qiao2018deep} are trained with 
the objective to maximize the agreement between them. DMT \cite{2020DMT} updates one network with the fixed pseudo labels from another iteratively, and dynamically re-weights 
the loss depending on the degree of the disagreement. To deal with the coupling issue, Robust Mutual Learning \cite{RML} generates pseudo labels from the mean-teachers, which 
transfers the supervision in an indirect manner, and introduces the self-rectification for the labels according to the internal knowledge. In addition, the cross-image label 
agreement and disagreement are leveraged to mine the cross-image semantic relations for weakly supervised semantic segmentation \cite{sun2020mining}.

Different from them, the two networks in CPCL are trained under the pseudo supervision of the `intersection' and `union' parts, achieving the collaboration of the conservative 
evolution and the progressive exploration, and the loss is re-weighted according to the prediction confidence to reduce the influences of the suspicious pseudo labels.

\section{Methodology}

\begin{figure*}[h]
  \centering
  \includegraphics[scale=0.55]{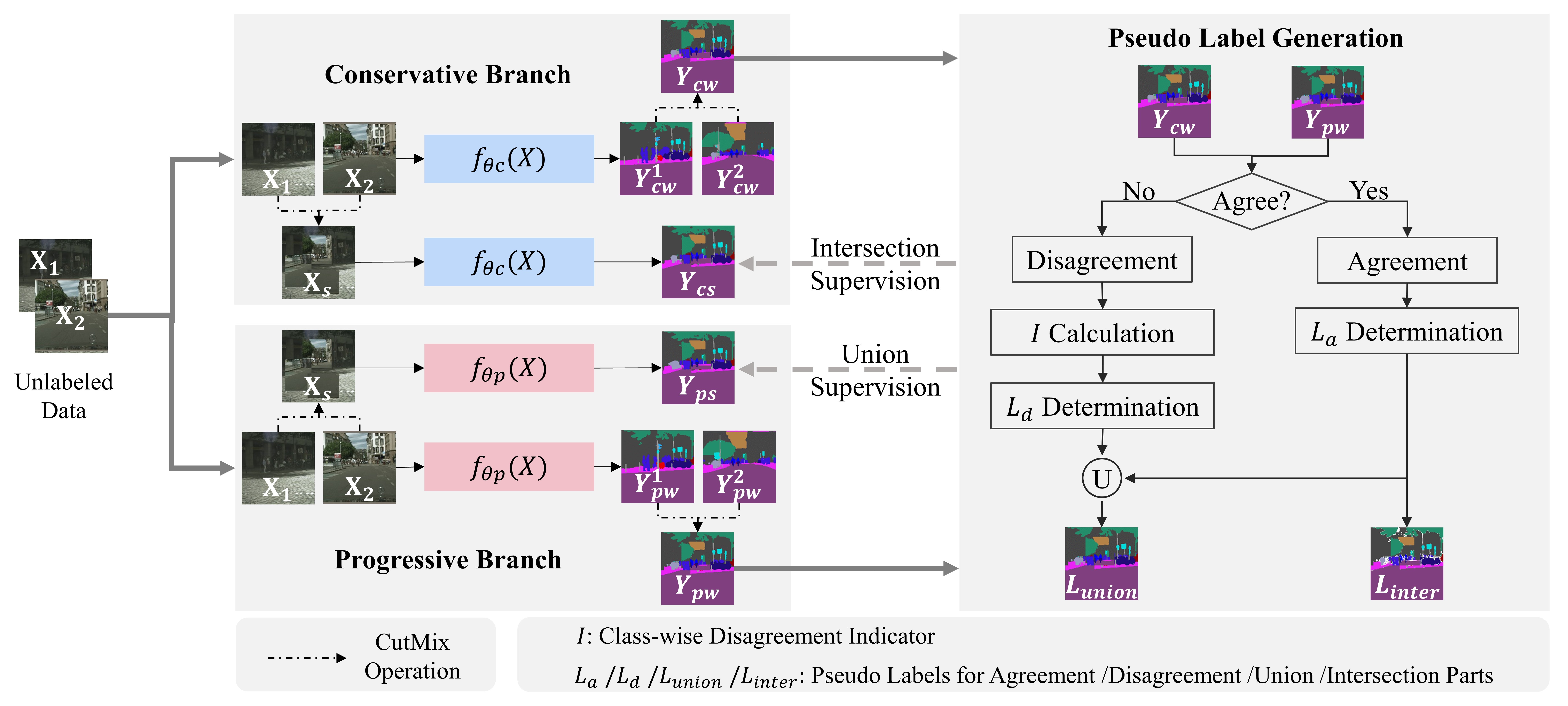}
  \caption{The framework of the proposed conservative-progressive collaborative learning (CPCL). Two networks are trained in parallel, which share the same structure but are 
  initialized differently. The pseudo supervision is generated on the basis of the agreement and disagreement instead of utilizing the output of the two branches directly. 
  The conservative branch is under the intersection pseudo supervision using the high-quality pseudo labels for evolution, while the progressive branch is supervised by 
  the union pseudo supervision utilizing the large quantity of pseudo labels for exploration. The two branches seek common ground while reserving differences, and achieve 
  the collaboration of the conservative and the progressive. The visualization results in the figure are output from the networks at the later training stage.}
  \label{fig:cpcl}
\end{figure*}

In this section, we firstly expound the problem definition, and then propose the conservative-progressive collaborative learning framework. To generate the pseudo labels for the progressive 
union supervision, a label determination approach based on class-wise disagreement indicator for disagreement parts is proposed. We further present the confidence-based dynamic 
loss on the basis of the prediction confidence to deal with the unreliable pseudo labels.

\subsection{Problem Definition}

The data provided in the SSL tasks including labeled data and unlabeled data, and the unlabeled training data set is much larger than the labeled one. Without loss of 
generality, let $D_l = \{(X_l^1, G^1), ... , (X_l^n, G^n), ... , (X_l^N, G^N)\}$ denote the $N$ labeled data, while the $M$ unlabeled data is represented as 
$D_u = \{X_u^1, ... , X_u^m, ... , X_u^M\}$. 
$X_l^n$ and $G^n$ are the labeled input image and the corresponding label for the semantic segmentation, and $X_u^m$ is the $m$-th unlabeled input image. Specifically, 
$X_l^n, X_u^m \in \mathbb{R}^{H \times W}$, and $G^n \in \mathbb{R}^{C \times H \times W}$. 

To perform well on the test data, the key problem is to leverage the large number of unlabeled data $D_u$ for the training of the segmentation network $f_{\theta}(X)$. In this 
work, the main architecture contains two segmentation networks, $f_{\theta_c}(X)$ and $f_{\theta_p}(X)$. While utilizing the labeled data $D_l$, the networks are trained in a traditional supervised manner, the proposed CPCL is introduced for the training on the 
unlabeled data $D_u$. 

\subsection{Conservative-Progressive Collaborative Learning}

To not only take the advantage of the high-quality labels but also make the full use of the large quantity of pseudo labels, we propose a novel semi-supervised learning 
approach for semantic segmentation, named conservative-progressive collaborative learning. It is composed of two paralleled branches, one is for conservative evolution, 
and the other is for progressive exploration. The conservative branch is under the intersection pseudo supervision to ensure a more reliable supervision by utilizing 
high-quality labels, while the progressive branch is supervised by the union pseudo labels to achieve the exploration of the disagreement parts. The framework of the 
proposed CPCL is shown in Fig.~\ref{fig:cpcl}.

The segmentation network in each branch shares the same structure and has different initialization. Two unlabeled images, $X_1$ and $X_2$, are input, and an augmented 
image $X_s$ is generated utilizing the strong augmentation methods, e.g., CutMix, CutOut and CowMix \cite{2019Semi}. CutMix is adopted in this paper, which requires two 
input images and selects the output pixel based on a random mask,
\begin{equation}
  mix(a, b, mask) = (1 - mask) \odot a + mask \odot b,
\end{equation}
where $a$ and $b$ are the two inputs. The augmented image $X_s$ is generated as
\begin{equation}
  X_s = mix(X_1, X_2, mask).
\end{equation}

Take the conservative branch as example, the three images are fed into the 
network $f_{\theta_c}(X)$, and the segmentation predictions are $Y_{cw}^1$, $Y_{cw}^2$, and $Y_{cs}$.

\begin{equation}
  Y_{cw}^1 \leftarrow \argmax_y f_{\theta_c}(y|X_1) ,
\end{equation}

\begin{equation}
  Y_{cw}^2 \leftarrow \argmax_y f_{\theta_c}(y|X_2) ,
\end{equation}

\begin{equation}
  Y_{cs} \leftarrow \argmax_y f_{\theta_c}(y|X_s) .
\end{equation}

The output $Y_{cw}^1$ and $Y_{cw}^2$ are further mixed using the same mix mask of the input, and the result is $Y_{cw}$,
\begin{equation}
  Y_{cw} = mix(Y_{cw}^1, Y_{cw}^2, mask).
\end{equation}

Similarly, we can get the corresponding output of the other 
branch, which are denoted as $Y_{pw}^1$, $Y_{pw}^2$, $Y_{ps}$, and the mixed output $Y_{pw}$. Instead of directly applying cross pseudo supervision between the predictions 
of the weak-augmented images and the strong-augmented images, e.g., $Y_{cw} \rightrightarrows Y_{ps}$ and $Y_{pw} \rightrightarrows Y_{cs}$, there are some additional operations applied 
to the $Y_{cw}$ and $Y_{pw}$ to generate the pseudo labels.

The pseudo label generation is on the basis of the agreement and disagreement, and both of them are pixel-wise. For a pixel $p_i$, if the prediction of $Y_{cw}$ is the 
same as that of $Y_{pw}$, the two networks achieve agreement at $p_i$, otherwise they disagree with each other.

\begin{eqnarray}
  agreement =  
  \begin{cases}
    True & y_{cw}^i = y_{pw}^i\\
    False & otherwise
  \end{cases} ,
\end{eqnarray}
where $y_{cw}^i$ is the prediction of $Y_{cw}$ at $p_i$, and $y_{pw}^i$ is the corresponding prediction of $Y_{pw}$. 

The pseudo label of a specific pixel is easy to determine, if the two predictions of the networks agree with each other. It is determined as the prediction at that pixel.

\begin{equation}
  l_a^i = y_{cw}^i, \; if \; y_{cw}^i = y_{pw}^i ,
\end{equation}
where $l_a^i$ is the pseudo label for the agreement parts $L_a$ at the pixel $p_i$. However, it is much more complex to determine the pixel-wise pseudo labels for the disagreement 
parts $L_d$, as the predictions of the two networks are different. Instead of selecting the prediction with higher confidence, we determine the pseudo label based on the 
class-wise disagreement indicator, which is elaborated detailedly in the next subsection. Obtaining the pseudo labels for both the agreement and disagreement parts, the pseudo labels 
for union $L_{union}$ is compounded utilizing $L_a$ and $L_d$,

\begin{equation}
  L_{union} = L_a \cup L_d .
\end{equation}
Much simpler than that, the pseudo labels for intersection $L_{inter}$ is directly generated using the $L_a$. 

Finally, the intersection and union pseudo supervision are applied to the conservative branch and progressive branch, respectively.

\begin{equation}
  L_{inter} \rightrightarrows Y_{cs} ,
\end{equation}
\begin{equation}
  L_{union} \rightrightarrows Y_{ps} ,
\end{equation}

The whole semi-supervised learning approach seeks common ground of the two paralleled branches while reserving differences, and achieves the collaboration of 
the conservative and the progressive.

Since the predictions of the two networks tend to agree with each other at the later training stage, which is indicated by the large overlap ratio (greater than 97\%), 
we only preserve the network in conservative branch to generate segmentation predictions in inference mode.

\subsection{Pseudo Label Determination for Disagreement}

To deal with the disagreement between two predictions at the same pixel, a pseudo label determination approach for the disagreement part is introduced in this subsection, 
which is based on the class-wise disagreement indicator.

\begin{figure}[h]
  \centering
  \includegraphics[scale=0.6]{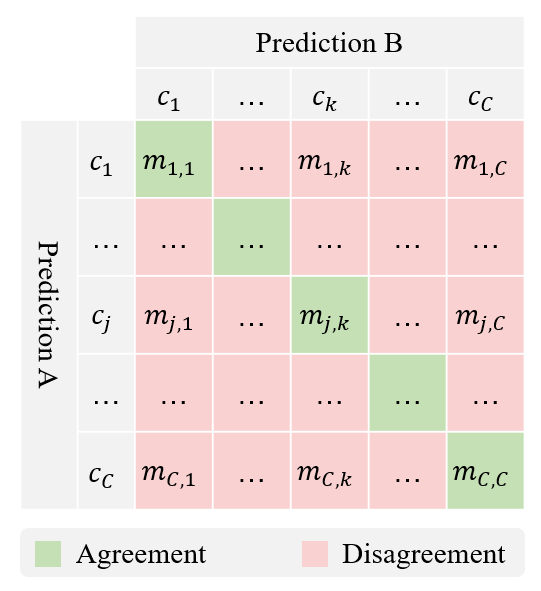}
  \caption{The Illustration of the agreement matrix. $m_{j,k}$ is the number of pixels where the $y_{cw}^i$ is $c_j$ and $y_{pw}^i$ is $c_k$ ($j,k \in [1,C]$), and $C$ is 
  the number of classes. Two predictions achieve agreement when $j=k$, and otherwise they disagree with each other.}
  \label{fig:AM}
\end{figure}

The typical and feasible approach is selecting the prediction with higher confidence as the pseudo label. However, this approach only focuses on the information of 
the exact pixel, which is limited. Addressing that, we turn to the class-wise disagreement indicator of the whole prediction to take a look from a macro perspective. 
Specifically, we construct the agreement matrix $M \in \mathbb{R}^{C \times C}$, as shown in Fig.~\ref{fig:AM}, and calculate the class-wise disagreement indicator 
$I$ based on that. The semantic class with a higher $I$ is the difficult class, which is more likely to cause confusion. The pseudo label at the pixel $p_i$, 
where $y_{cw}^i \neq y_{pw}^i$, is determined as the prediction with higher disagreement indicator to achieve the progressive exploration for the disagreement part 
with strong curiosity. The pseudo label determination algorithm based on class-wise disagreement indicator is shown in Algorithm~\ref{alg:PLDA}.

\begin{algorithm}[ht]
  \caption{Class-wise Disagreement Indicator Based Pseudo Label Determination Algorithm}
  \label{alg:PLDA}
  \begin{algorithmic}[1]
    \Statex
    \Require
      The prediction output from the conservative branch $Y_{cw}$ and that output from the progressive branch $Y_{pw}$;
      \Statex

    \Ensure
      The pseudo labels for the disagreement parts $L_d$;
    \Statex 

    \State Calculating the agreement matrix $M \in \mathbb{R}^{C \times C}$, where $C$ is the number of classes;
    \Statex 
    \Statex $m_{j,k} \gets$ the number of pixels where the $y_{cw}^i$ is $c_j$ and $y_{pw}^i$ is $c_k$, and $j,k \in [1,C]$.
    \Statex 

    \State Calculating the class-wise disagreement indicator $I$;
    \Statex 
    \Statex \centerline{$I_j = 2 - \frac{m_{j,j}}{\sum_{k=1}^C m_{j,k}} - \frac{m_{j,j}}{\sum_{k=1}^C m_{k,j}}$}
    \Statex \centerline{$I_k = 2 - \frac{m_{k,k}}{\sum_{k=1}^C m_{j,k}} - \frac{m_{k,k}}{\sum_{k=1}^C m_{k,j}}$}
    \Statex where $j,k \in [1, C]$ are the index of the class.
    \Statex 

    \State Determination of the pseudo label at pixel $p_i$;
    \Statex 
    \begin{eqnarray*}
      l_d^i = 
      \begin{cases}
        c_j & I_j \ge I_k, j \neq k\\
        c_k & I_k \ge I_j, j \neq k
      \end{cases}
    \end{eqnarray*}

    \leftline{\Return $L_d$.}
  \end{algorithmic}
\end{algorithm}

\begin{figure*}
  \centering
  \includegraphics[scale=0.1]{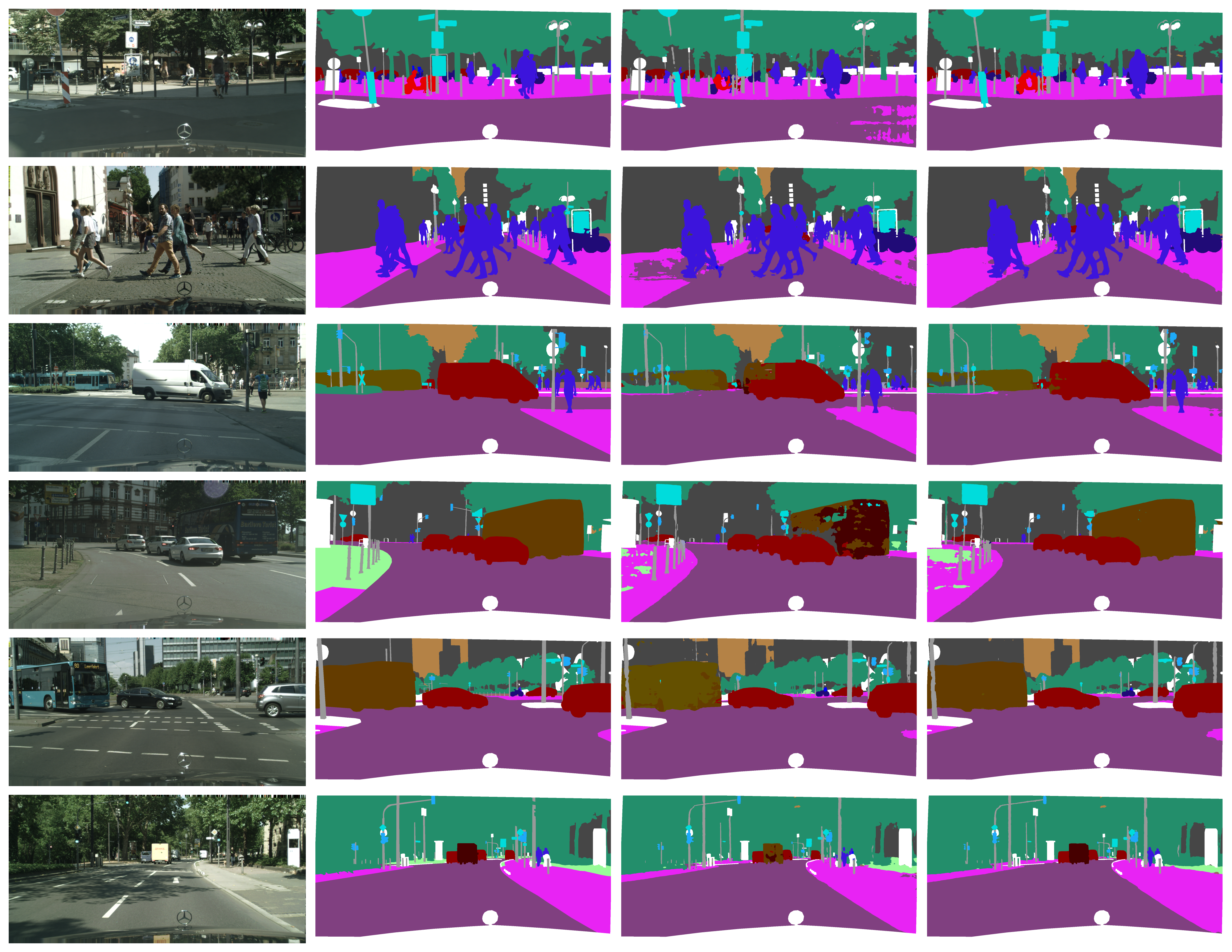}
  \includegraphics[scale=0.5]{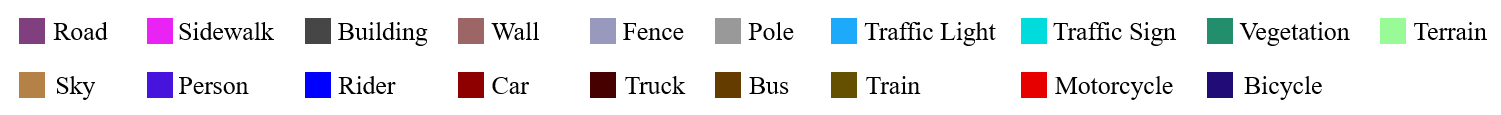}
  \caption{Visualization examples on Cityscapes validation set (1/8 Labels). \textbf{Left}: Input; \textbf{Middle left}: Groundtruth; \textbf{Middle right}: Predictions of the 
  supervised-only method; \textbf{Right}: Predictions of CPCL.}
  \label{fig:cs_demo}
\end{figure*}

\subsection{Confidence-based Dynamic Loss}

Benefit from the approaches proposed in the previous subsections, the pseudo labels of the unlabeled data are obtained. However, the noisy label is ineluctable, 
and there always be some errors in the generated pseudo labels since it is automatically annotated by the algorithm instead of human. Many researches have explored 
image denoising algorithms \cite{5504216, 7577781}. Different from the image noise, the pseudo label noise is difficult to be either modeled or denoised. Addressing that, the 
confidence-based dynamic loss is proposed in this subsection to deal with the unreliable pseudo labels. Contrary to the confidence-thresholding, which requires a 
predefined threshold and could not take the full advantage of all the unlabeled data, we utilize the confidence for loss re-weighting. 

The pixel-wise prediction confidence is defined as the maximum softmax probability. Let $b_c^i$ denotes the prediction confidence of the conservative branch at the $i$-th 
pixel, similarly, that of the progressive branch is denoted as $b_p^i$. Therefore, the dynamic confidence-based weight is defined as:

\begin{eqnarray}
  \omega_u^i =  
  \begin{cases}
    \frac{1}{2}(b_c^i + b_p^i) & y_{cw}^i = y_{pw}^i\\
    b_c^i & y_{cw}^i \neq y_{pw}^i, l_d^i \leftarrow y_{cw}^i\\
    b_p^i & y_{cw}^i \neq y_{pw}^i, l_d^i \leftarrow y_{pw}^i
  \end{cases} .
\end{eqnarray}

Thus, the influence of the suspicious pseudo label with low confidence is reduced. The re-weighted loss on unlabeled data $\mathcal{L_U}$ is then defined as

\begin{equation}
  \begin{small}
  \begin{aligned}
  \mathcal{L_U} 
&= \frac{1}{W \times H} \sum_i^{W \times H} \omega_u^i \CE(l_{inter}^i, f_{\theta_c}^i(X_s)) \\
&+ \frac{1}{W \times H} \sum_i^{W \times H} \omega_u^i \CE(l_{union}^i, f_{\theta_p}^i(X_s)) ,      
  \end{aligned}
  \end{small}
\end{equation}
where $ l_{inter}^i $ and $ l_{union}^i $ are the pseudo labels of $L_{inter}$ and $L_{union}$ at the $i$-th pixel, the $f_{\theta}^i(X_s)$ is the output of the 
segmentation network at the $i$-th pixel for the unlabeled image $X_s$, and $CE(\cdot)$ is the cross-entropy loss function. The loss is re-weighted at each pixel 
according to the dynamic confidence-based weight.

The training leveraging the labeled data is in a traditional supervised manner.
\begin{equation}
  \mathcal{L_S} = \CE(G, f_{\theta}(X_l)) ,
\end{equation}
where $X_l \in D_l$ is the labeled image, and $G$ is the corresponding ground truth. The separated loss for supervised and unsupervised training is combined to 
the whole training objective, as
\begin{equation}
  \mathcal{L} = \mathcal{L_S} + \gamma \mathcal{L_U} ,
\end{equation}
where $\gamma$ is the trade-off weight to balance the two parts.

\section{Experiments}
In this section, we evaluate the performance of the proposed approach CPCL on two public datasets, Cityscapes \cite{Cityscapes} and PASCAL VOC 2012 \cite{pascalvoc}. 
The experiments are implemented in the PyTorch on a server with a NVIDIA V100.

\subsection{Implementation Detail and Dataset}
The segmentation networks share the same structure, DeepLabv3+, and the weights of the segmentation heads are initialized randomly. The initial learning rate is 
set as $10^{-4}$ and $5 \times 10^{-3}$ when training with Cityscapes and PASCAL VOC, respectively, and the learning rate is multiplied by $(1-\frac{iter}{max\_iter})^{0.9}$ 
in every iteration. The SGD with momentum is adopted. The momentum is fixed as $0.9$, and the weight decay is $10^{-4}$. Due to the limitation of the GPU, 
the batch size for Cityscapes is set as 3, and that for PASCAL VOC is 8. The loss weight $\gamma$ is 5 and 1 respectively. We follow the partition protocols of GCT \cite{GCT}, which randomly extract 1/2, 1/4, 1/8, and 1/16 
of the whole training set as the labeled subset, and regard the rest of the data as the unlabeled subset. If strong augmentation is to be applied, three rectangle 
regions of random ratio (with the range of $[0.25, 0.5]$) to the input is positioned randomly and augmented with CutMix strategy. 

\begin{table*}
  \footnotesize
  \caption{Quantitative results of the improvements over the supervised baseline on Cityscapes. All scores are in \%, and the performance improvement is shown 
  in brackets. (\textbf{Flat}: road and sidewalk; \textbf{Human}: person and rider; \textbf{Vehicle}: car, truck, bus, train, motorcycle, and bicycle; \textbf{Construction}: 
  building, wall, and fence; \textbf{Object}: pole, traffic lights, and traffic sign; \textbf{Nature}: vegetation and terrain; \textbf{Sky}: sky.)}
  \begin{center}
  \renewcommand{\arraystretch}{1.4}
  \begin{tabular}{cc|c|ccccccc} 
  \hline
    Partition   & Methods      & mIoU              & Flat              & Human              & Vehicle             & Construction             & Object             & Nature             & Sky      \\ \hline
    1/2         & Supervised   & 75.36             & 89.87             & 71.71              & 74.21               & 67.35                    & 69.91              & 78.52              & 94.52    \\
    (1488)      & CPCL         & 78.17 (\bf 2.81)  & 91.44 (\bf 1.57)  & 73.88 (\bf 2.16)   & 79.29 (\bf 5.08)    & 69.15 (\bf 1.80)         & 72.44 (\bf 2.53)   & 79.68 (\bf 1.16)   & 94.72 (\bf 0.20)   \\ \hline
    1/4         & Supervised   & 73.22             & 88.88             & 71.53              & 70.40               & 64.07                    & 69.80              & 75.98              & 94.34    \\
    (744)       & CPCL         & 76.98 (\bf 3.76)  & 90.85 (\bf 1.96)  & 72.77 (\bf 1.24)   & 77.40 (\bf 7.00)    & 68.24 (\bf 4.17)         & 71.89 (\bf 2.09)   & 78.06 (\bf 2.08)   & 94.50 (\bf 0.17)   \\ \hline
    1/8         & Supervised   & 68.63             & 87.89             & 69.81              & 60.02               & 60.68                    & 67.85              & 74.43              & 94.01    \\
    (372)       & CPCL         & 74.60 (\bf 5.97)  & 89.48 (\bf 1.59)  & 72.88 (\bf 3.07)   & 72.89 (\bf 12.88)   & 65.05 (\bf 4.36)         & 70.49 (\bf 2.64)   & 77.13 (\bf 2.70)   & 94.40 (\bf 0.39)   \\ \hline
    1/16        & Supervised   & 61.67             & 85.27             & 65.66              & 46.88               & 53.49                    & 63.03              & 72.71              & 93.51    \\
    (186)       & CPCL         & 69.92 (\bf 8.25)  & 88.41 (\bf 3.15)  & 70.15 (\bf 4.49)   & 61.96 (\bf 15.08)   & 62.69 (\bf 9.20)         & 68.50 (\bf 5.47)   & 75.97 (\bf 3.26)   & 94.02 (\bf 0.51)   \\ \hline

  \end{tabular}
  \end{center}
  \label{tab:cs_sup_unsup}
\end{table*}

\begin{table*}[h]
 
  \footnotesize
  \caption{Quantitative results of the improvements over the supervised baseline on PASCAL VOC 2012. All scores are in \%, and the performance improvement is shown 
  in brackets. (\textbf{Animal}: bird, cat, cow, dog, horse, and sheep; \textbf{Vehicle}: aeroplane, bicycle, boat, bus, car, motorbike, and train; 
  \textbf{Indoor}: bottle, chair, dining table, potted plant, sofa, and monitor;\textbf{Person}: person;  \textbf{Background}: background.)}
  \begin{center}
  \renewcommand{\arraystretch}{1.3}
  \begin{tabular}{cc|c|ccccc} 
  \hline
    Partition   & Methods      & mIoU              & Animal             & Vehicle             & Indoor                   & Person            & Background       \\ \hline
    1/2         & Supervised   & 74.05             & 84.45              & 76.90               & 55.76                    & 81.95             & 93.57            \\
    (5291)      & CPCL         & 75.30 (\bf 1.25)  & 86.63 (\bf 2.18)   & 77.23 (\bf 0.33)    & 57.11 (\bf 1.35)         & 84.23 (\bf 2.28)  & 94.12 (\bf 0.54) \\ \hline
    1/4         & Supervised   & 71.66             & 78.98              & 74.31               & 55.91                    & 82.05             & 93.23            \\
    (2646)      & CPCL         & 74.58 (\bf 2.92)  & 85.39 (\bf 6.41)   & 76.62 (\bf 2.31)    & 56.69 (\bf 0.78)         & 83.65 (\bf 1.59)  & 93.70 (\bf 0.47) \\ \hline
    1/8         & Supervised   & 67.16             & 73.47              & 71.94               & 48.66                    & 81.98             & 92.28            \\
    (1323)      & CPCL         & 73.74 (\bf 6.58)  & 84.24 (\bf 10.77)  & 76.53 (\bf 4.59)    & 54.91 (\bf 6.26)         & 84.36 (\bf 2.38)  & 93.60 (\bf 1.33) \\ \hline
    1/16        & Supervised   & 62.00             & 67.65              & 67.88               & 41.56                    & 79.99             & 91.63            \\
    (662)       & CPCL         & 71.66 (\bf 9.66)  & 82.73 (\bf 15.08)  & 76.24 (\bf 8.36)    & 49.70 (\bf 8.14)         & 83.57 (\bf 3.58)  & 93.02 (\bf 1.39) \\ \hline

  \end{tabular}
  \end{center}
  \label{tab:voc_sup_unsup}
\end{table*}
	
\textbf{Cityscapes} is a dataset mainly designed for the urban driving scenes, which consists of 2975 images for training and 500 images for validation. Each image 
has a resolution of $2048 \times 1024$, and we random crop it to $800 \times 800$ with the original resolution kept. Each pixel is annotated with 
one of the semantic labels from 19 classes. 

\textbf{PASCAL VOC 2012} is a standard semantic segmentation dataset for common objects, which comprises 20 foreground classes along with a background class. The 
original dataset has 1464 images for training and 1449 validation images. We follow the previous works to generate the augmented version, which is combined with 
the extra annotation set from the Segmentation Boundaries Dataset (SBD) \cite{SBD}. Thus, the full training set consists of 10582 images. We keep the original resolution and 
random crop the images to $512 \times 512$.

\subsection{Effectiveness Study}

Firstly, we evaluate our approach on Cityscapes to verify the effectiveness for the complex street scenes, and report the results on the validation set (500 images) for 
all partition protocols. The visualization examples on Cityscapes validation set are shown in Fig.~\ref{fig:cs_demo}. It can be seen from the first and the second 
rows that some pixels of road and sidewalk are missegmented by the supervised-only method. There are also confusion among train, truck and bus, as shown in the next 
four rows of the third column. Benefit from the proposed approach CPCL, the errors are corrected, and the predictions are more accurate compared with the supervised-only one.

\begin{figure}[h]
  \centering
  \includegraphics[scale=0.33]{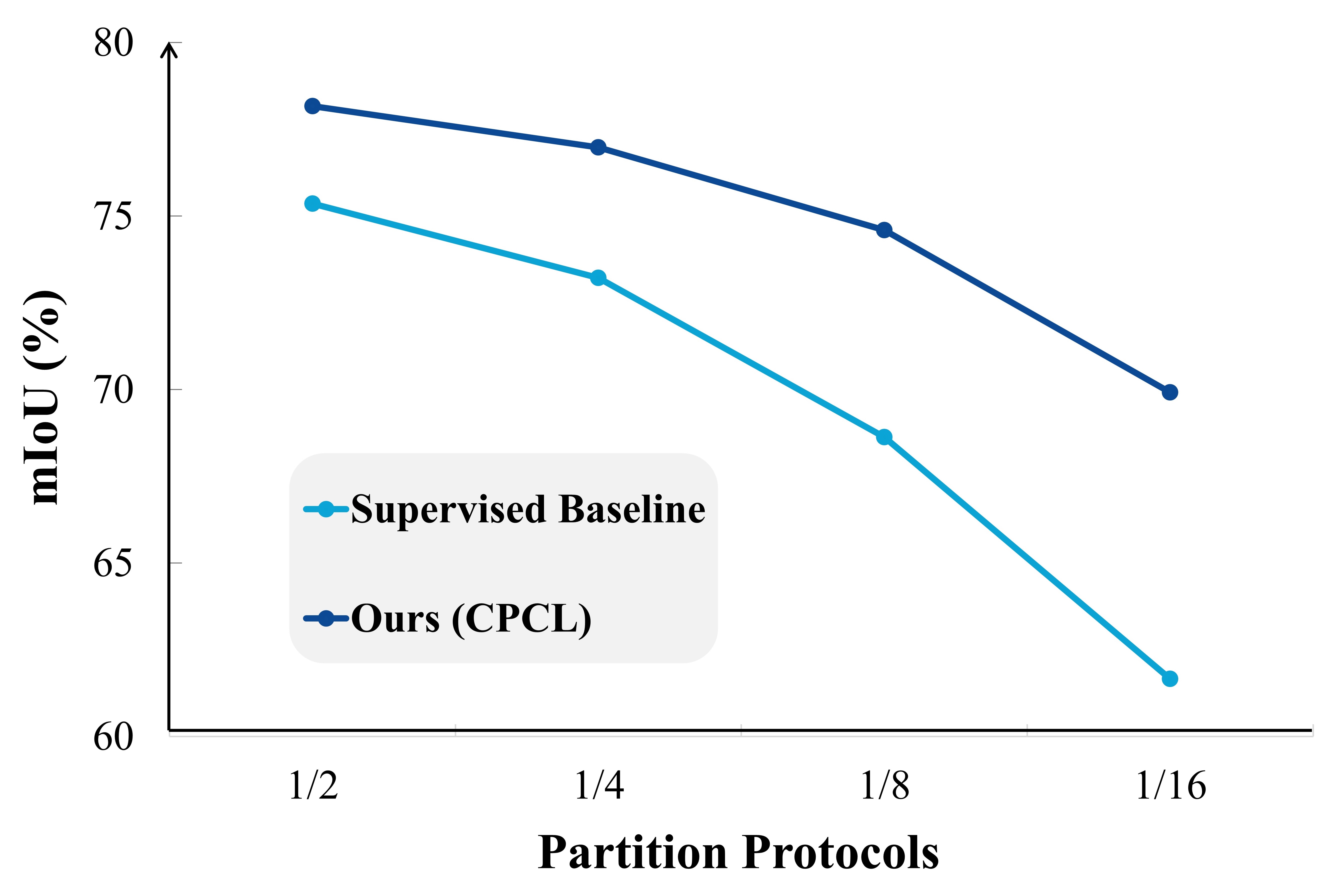}
  \caption{Improvements over the supervised baseline on Cityscapes.}
  \label{fig:city_sup_unsup}
\end{figure}

The improvements over the supervised baseline on Cityscapes are significant, as shown in Fig.~\ref{fig:city_sup_unsup}. Our approach consistently outperforms the 
DeepLabv3+ model under all partition protocols, and the gain of the CPCL increases as the ratio of the labeled data decreases. To further study the improvements, 
we report the quantitative results in Table~\ref{tab:cs_sup_unsup}, which is more detailed. The performance increments are shown in the brackets. It can be seen 
that the gains of the approach are 2.81\%, 3.76\%, 5.97\%, and 8.25\% under 1/2, 1/4, 1/8, and 1/16 partition protocols separately. The utilization of the unlabeled 
data let the model achieve the similar performance as the one leverages more labeled data but only under supervised learning, e.g., the performance of CPCL under 
1/16 partition is 69.92\% and that of the supervised-only method under 1/8 partition is 68.63\%. Compared with other class groups, the improvement for the vehicle 
is obvious, which achieves 15.08\% increasement when only 186 labeled images are used. The gain on the sky class is limited since the supervised-only model has already achieved 
fine performance (about 94\%) on that semantic class.

\begin{figure}[h]
  \centering
  \includegraphics[scale=0.33]{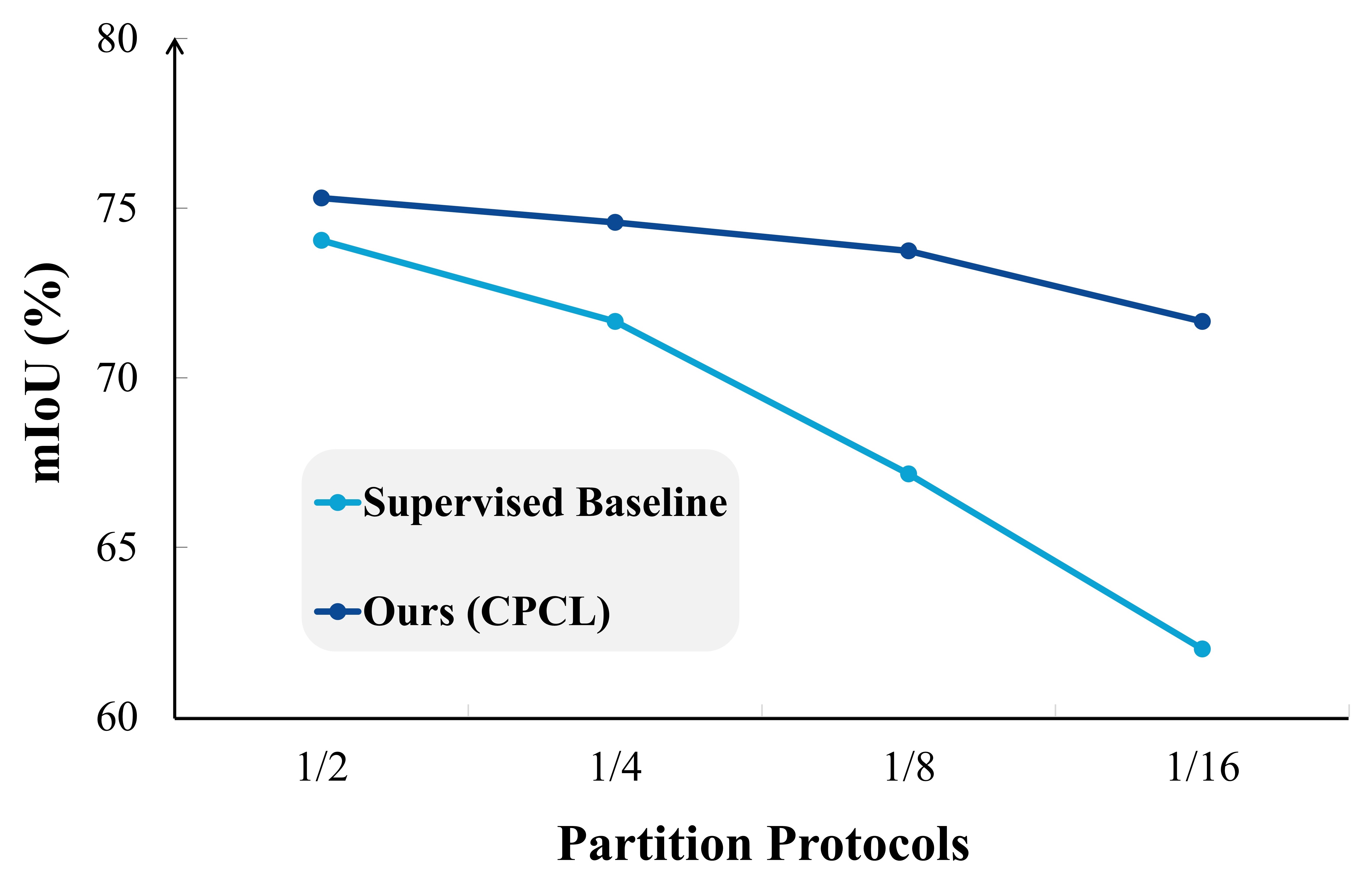}
  \caption{Improvements over the supervised baseline on PASCAL VOC 2012.}
  \label{fig:voc_sup_unsup}
\end{figure}

Different from Cityscapes, which focuses on traffic scenes, PASCAL VOC 2012 has richer scenes, and is the most commonly used benchmark for semi-supervised semantic 
segmentation. We evaluate our approach on it, and report the results on the validation set (1456 images) for all partition protocols.

In line with expectations, the CPCL consistently outperforms the supervised-only method under all partition protocols, and the gain is more obvious when fewer labeled 
images are used, as shown in Fig.~\ref{fig:voc_sup_unsup}. It can be seen from Table~\ref{tab:voc_sup_unsup} that all of the metrics are increased due to the 
utilization of the unlabeled data. For different partitions, our approach achieves a performance increase of 2.81\% (under 1/2 labels) to 8.25\% (under 1/16 labels) 
over the baseline. The performance improvement of the animal class group is the most significant among the five groups, which goes up to 15.08\% when only using 662 labeled 
images.

\subsection{Ablation Study}

The effectiveness of the proposed modules is further evaluated on Cityscapes under 1/8 partition protocol. We report the quantitative results in Table~\ref{tab:ablation}. 
The intersection supervision and the union supervision are evaluated individually by training both the two paralleled networks under each of them. The cross-entropy loss 
function without re-weighting is used when the confidence-based dynamic loss is not adopted.

\begin{table}[h]
 
  \footnotesize
  \caption{Quantitative results of the ablation study on Cityscapes under 1/8 partition protocol.}
  \begin{center}
  \renewcommand{\arraystretch}{1.3}
  \begin{tabular}{ccc|c} 
  \hline
    \multicolumn{2}{c}{Pseudo Supervision}             & \multirow{2}*{Dynamic Loss} & \multirow{2}*{mIoU (\%)} \\
    \cline{1-2}Intersection  & Union                   &                             &                          \\ \hline
    \checkmark               &                         &                             & 71.57                    \\
                             & \checkmark              &                             & 72.45                    \\
    \checkmark               & \checkmark              &                             & 73.58                         \\
    \checkmark               &                         & \checkmark                  & 73.94                    \\
                             & \checkmark              & \checkmark                  & 74.15                    \\ \hline
    \checkmark               & \checkmark              & \checkmark                  & 74.60                    \\ \hline

  \end{tabular}
  \end{center}
  \label{tab:ablation}
\end{table}

It can be seen that all of the three proposed modules are effective, resulting in the performance of the CPCL. First of all, the mIoU is improved from 68.63\% to 71.57\% 
when the intersection pseudo supervision is adopted. The improvement of 2.94\% is achieved by leveraging a large number of unlabeled images. Since the intersection 
supervision only focuses on the agreement parts, the supervision is conservative and the improvement is limited. Different from that, the union pseudo supervision explores 
the disagreement parts with curiosity and leads to a more obvious improvement of 3.82\% (from 68.63\% to 72.45\%). These two pseudo supervisions act in two different ways, 
conservative and progressive, and the collaboration of them improves the mIoU to 73.58\%. Addressing the noisy label, we propose the confidence-based dynamic loss. The mIoU is 
increased by 2.37\% and 1.70\% as the dynamic loss is introduced. The improvement of the intersection supervision is more obvious, because the determination of the pseudo 
labels for the agreement parts is pixel-level, which is more likely influenced by the noisy label. Benefit from both the proposed pseudo supervision and the proposed 
dynamic loss, the mIoU of CPCL achieves 74.6\%, overperforming the supervised baseline with 5.97\%.

\begin{figure}[h]
  \centering
  \includegraphics[scale=0.28]{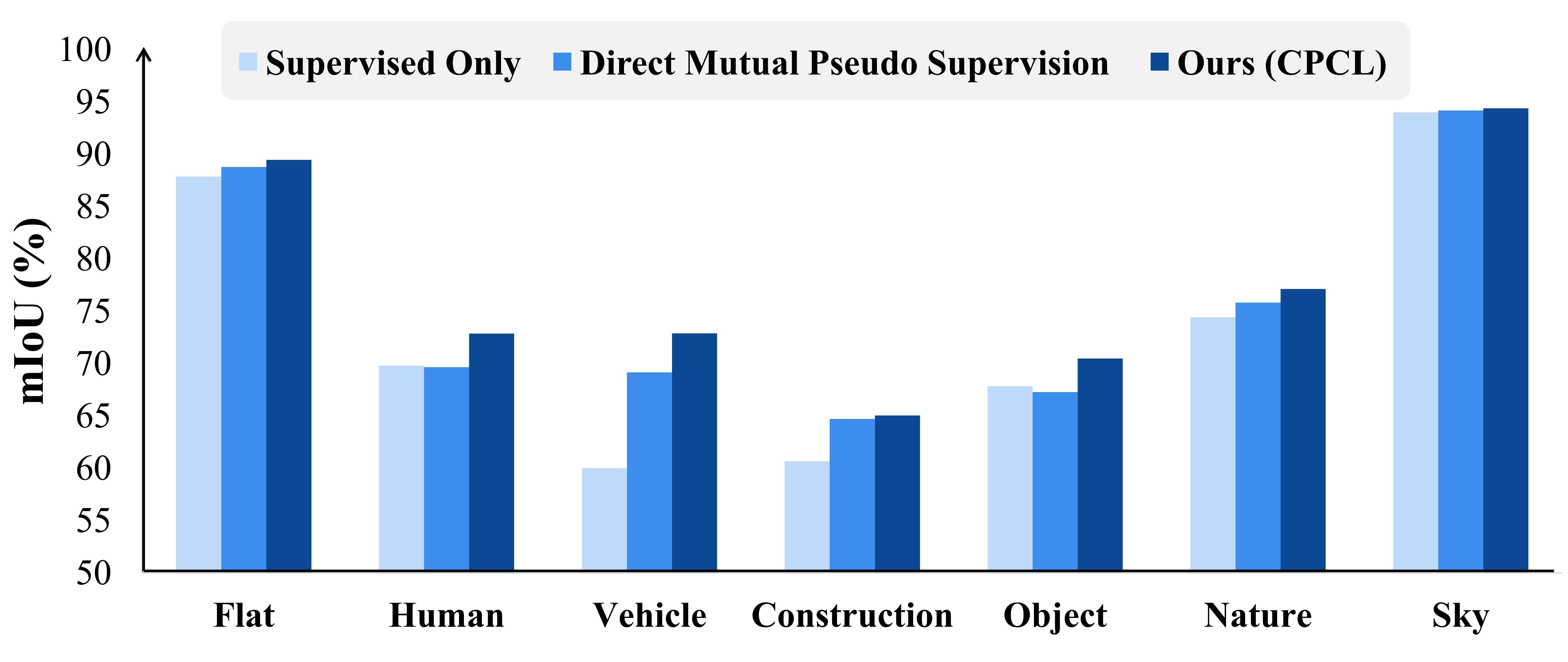}
  \caption{Effectiveness of the intersection-union supervision.}
  \label{fig:com_ablation}
\end{figure}

The proposed method is further compared with the mutual pseudo supervision, which directly utilizes the predictions to generate the pseudo labels and 
applies pseudo supervision based on that. The performances on the seven class groups are shown in Fig.~\ref{fig:com_ablation}. It's worth noting that the 
performance of the direct mutual pseudo supervision is slightly worse than that of the supervised-only method on human and object groups. Those two 
groups consist of person, rider, traffic light, and traffic sign, which are small sized and complex. Thus, the influence of the erroneous pseudo supervision is 
obvious. Benefit from our pseudo labels generation method, the pseudo supervision is more reliable, resulting in the performance improvements on all the groups.

\begin{table}[h]
 
  \footnotesize
  \caption{Comparison of four pseudo labeling strategies for the disagreement part on Cityscapes under 1/8 partition protocol.}
  \begin{center}
  \renewcommand{\arraystretch}{1.3}
  \begin{tabular}{cc|c} 
  \hline

    \multicolumn{2}{c|}{Pseudo Labeling Strategy}              & mIoU (\%) \\ \hline
    \multirow{2}*{pixel-wise confidence}  &  lower             &  74.30    \\
                                          &  higher            &  74.25    \\ \hline
    \multirow{2}*{class-wise confusion}   &  lower             &  74.17    \\
                                          &  higher (ours)     &  74.60    \\ \hline
    
  \end{tabular}
  \end{center}
  \label{tab:labeling-strategy}
\end{table}

Finally, we study different pseudo labeling strategies for the disagreement part and analyze the quantitative results. Four related strategies are compared, 
including the pseudo labeling on the basis of the pixel-wise confidence (higher and lower) and class-wise confusion (higher and lower), and the results are 
reported in Table~\ref{tab:labeling-strategy}. The pixel-wise confidence indicates the uncertainty of the exact pixel. Pseudo labeling with higher confidence 
is a common conservative strategy. However, the improvement is less obvious when collaborating with the intersection pseudo supervision, for both of them act 
conservatively. Exploration on the basis of lower confidence is progressive, but it is more likely influenced by pixel-wise noise. The 
proposed dynamic loss can reduce the influence to some extent, and the mIoU of it achieves 74.30\%. Different from the pixel-wise confidence, the class-wise 
confusion can measure the semantic ambiguity from a macro perspective. Pseudo labeling with lower confusion is similar to that based on higher confidence, 
but at class-level, which has the worst performance (74.17\% mIoU) among the four compared strategies. Since exploring the difficult semantic class with 
strong curiosity may help the models out the local optimization, the proposed strategy used in CPCL achieves a better performance.

\subsection{Comparison with Others}

We compare our approach with other semi-supervised segmentation methods in this subsection. All compared methods are based on DeepLabv3+ with ResNet-50 or ResNet-101, and the 
reported results of others are taken from \cite{CPS}.

\begin{table}[ht]
  \footnotesize
  \caption{Quantitative results of different methods on Cityscapes. The metric is mIoU(\%), and all compared methods are based on DeepLabv3+ with ResNet-50 or ResNet-101.}
  \begin{center}
  \renewcommand{\arraystretch}{1.3}
  \begin{tabular}{rccccc} 
  \hline
    Methods           & Network      & 1/16        & 1/8        & 1/4         & 1/2  \\ \hline
    Supervised        & ResNet-50    & 61.67       & 68.63      & 73.22       & 75.36      \\
    GCT \cite{GCT}    & ResNet-50    & 65.81       & 71.33      & 75.30       & 77.09      \\
                      & ResNet-101   & 66.90       & 72.96      & 76.45       & 78.58      \\
    MT \cite{MT}      & ResNet-50    & 66.14       & 72.03      & 74.47       & 77.43      \\
                      & ResNet-101   & 68.08       & 73.71      & 76.53       & \bf 78.59      \\      
    CCT \cite{CCT}    & ResNet-50    & 66.35       & 72.46      & 75.68       & 76.78      \\ 
                      & ResNet-101   & 69.64       & 74.48      & 76.35       & 78.29      \\ \hline
    CPCL (Ours)       & ResNet-50    & \bf 69.92   & \bf 74.60  & \bf 76.98   & 78.17      \\ \hline
   
  \end{tabular}
  \end{center}
  \label{tab:cs_sota}
\end{table}

Firstly, the quantitative results on Cityscapes are compared. As shown in Table~\ref{tab:cs_sota}, CPCL outperforms all reported methods with ResNet-50, and achieves similar 
or even better performance when compared with others with ResNet-101. In addition, it can be seen that the performance advantage becomes more evident when leveraging 
fewer labeled images for training.

Then, the comparison between the CPCL and others on PASCAL VOC 2012 is reported in Table~\ref{tab:voc_sota}. Due to the limitation of the GPU, the batch size is set as 7 when 
evaluating the CPCL with ResNet-101. Similar to that on Cityscapes, the performance advantage of our approach is more evident when using fewer labeled images. As the ratio of 
labeled data increases, the advantage decreases. However, CPCL with ResNet-101 still achieves state-of-the-art performance and outperforms the compared methods. Moreover, 
the CPCL with ResNet-50 even performs better than those with ResNet-101 under the partitions 1/8 labels.

\begin{table}[h]
  \footnotesize
  \caption{Quantitative results of different methods on PASCAL VOC 2012. The metric is mIoU(\%), and all compared methods are based on DeepLabv3+ with ResNet-50 or ResNet-101.
  Due to the limitation of the GPU, the batch size is set as 8 and 7, when evaluating the CPCL with ResNet-50 and ResNet-101.}
  \begin{center}
  \renewcommand{\arraystretch}{1.3}
  \begin{tabular}{rccccc} 
  \hline
    Methods                    & Network      & 1/16        & 1/8        & 1/4         & 1/2  \\ \hline
    Supervised                 & ResNet-50    & 62.00       & 67.16      & 71.66       & 74.05      \\
    GCT \cite{GCT}             & ResNet-50    & 64.05       & 70.47      & 73.45       & 75.20      \\
                               & ResNet-101   & 69.77       & 73.30      & 75.25       & 77.14      \\
    CCT \cite{CCT}             & ResNet-50    & 65.22       & 70.87      & 73.43       & 74.75      \\
                               & ResNet-101   & 67.94       & 73.00      & 76.17       & 77.56      \\
    CutMix-Seg \cite{2019Semi} & ResNet-50    & 68.90       & 70.70      & 72.46       & 74.49      \\
                               & ResNet-101   & 72.56       & 72.69      & 74.25       & 75.89      \\       
    MT \cite{MT}               & ResNet-50    & 66.77       & 70.78      & 73.22       & 75.41      \\ 
                               & ResNet-101   & 70.59       & 73.20      & 76.62       & 77.61      \\ \hline
    CPCL (Ours)                & ResNet-50    & 71.66       & 73.74      & 74.58       & 75.30      \\
                               & ResNet-101   & \bf 73.44   & \bf 76.40  & \bf 77.16   & \bf 77.67  \\ \hline           
   
  \end{tabular}
  \end{center}
  \label{tab:voc_sota}
\end{table}

In addition, the proposed framework CPCL does not restrict the backbones. The two parallel branches can use two different backbones. For example, one with 
ResNet-50 and another with ResNet-101. The experiments are further conducted on PASCAL VOC under 1/8 partition protocol. If the backbone of the conservative 
branch is changed to ResNet-101, the mIoU is improved to 73.94\%. The improvement is more obvious when ResNet-101 is used in the progressive branch, which 
achieves 74.85\% mIoU. Note that the reported result is the better one of the two parallel branches, and it is usually output from the progressive branch.

\begin{figure}[h]
  \centering
  \includegraphics[scale=0.35]{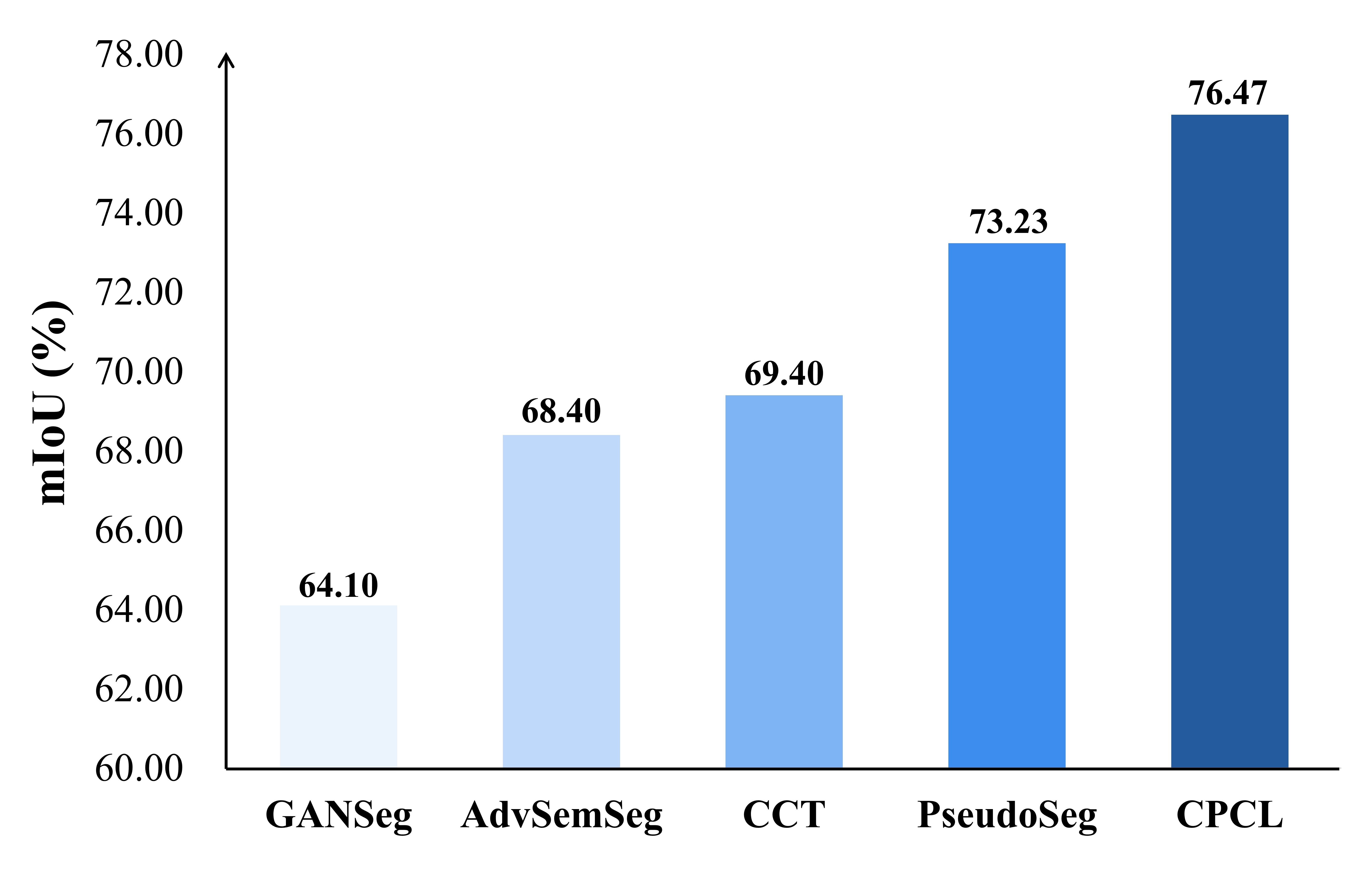}
  \caption{Comparison for full-supervision on PASCAL VOC 2012. We use the full PASCAL VOC 2012 training set (1.4k) as the labeled data, and the images from SBD (9k) as the 
  additional unlabeled data.}
  \label{fig:voc_full}
\end{figure}

\subsection{Improving Full-supervision with Additional Data}

The improvement over the full-supervision of our approach is verified in this subsection. The full PASCAL VOC 2012 training set (1.4k) is used as the labeled data, and the 
images from SBD (9k) are utilized as the additional unlabeled data. We compare the proposed approach with recent state-of-the-arts methods, as shown in Fig.~\ref{fig:voc_full}. 

All the results of others are taken from \cite{pseudoseg}. The backbone of GANSeg \cite{souly2017semi} is VGG16, while that of the CCT \cite{CCT} is ResNet-50, which is the same 
as ours. Both the AdvSemSeg \cite{ALS4} and the PseudoSeg \cite{pseudoseg} use ResNet-101 as the backbone. It can be seen that CPCL achieves the best performance among all 
the compared methods. The mIoU of our approach is 76.47\%, which is better than the GANSeg with a large margin (about 12\%). The second higher method, PseudoSeg, is still 3.24\% lower than ours. 

\subsection{Improving Few-supervision in Low-data Regime}

We further evaluate our approach in low-data regime on PASCAL VOC 2012. The partition protocols in PseudoSeg \cite{pseudoseg} is adopted in experiment, in which the labeled 
data set is constructed by sampling part of the images in the standard training set. The unlabeled set is generated by utilizing the additional unlabeled data in the SBD 
and the remaining images in the standard training set. The comparison between CPCL and other methods is reported in Table~\ref{tab:voc_few}.

\begin{table}[th]
  \footnotesize
  \caption{Quantitative results of different methods on PASCAL VOC 2012 in low-data regime. The metric is mIoU(\%), and the results of the compared methods are from \cite{pseudoseg}.
  }
  \begin{center}
  \renewcommand{\arraystretch}{1.3}
  \begin{tabular}{rccccc} 
  \hline
                               &              & \multicolumn{4}{c}{Exact Numbers of Labeled Data}          \\
    Methods                    & Network      & 92  & 183  & 366   & 732  \\ \hline
    CCT \cite{CCT}             & ResNet-50    & 33.10       & 47.60      & 58.80       & 62.10      \\
    VAT \cite{VAT}             & ResNet-101   & 36.92       & 49.35      & 56.88       & 63.34      \\
    AdevSemSeg \cite{ALS4}     & ResNet-101   & 39.69       & 47.58      & 59.97       & 65.27      \\
    GCT \cite{GCT}             & ResNet-101   & 46.04       & 54.98      & 64.71       & 70.67      \\
    MT \cite{MT}               & ResNet-101   & 48.70       & 55.81      & 63.01       & 69.16      \\                       
    CutMix-Seg \cite{2019Semi} & ResNet-101   & 55.58       & 63.20      & 68.36       & 69.84      \\
    PseudoSeg \cite{pseudoseg} & ResNet-101   & 57.60       & 65.50      & 69.14       & 72.41      \\ \hline

    CPCL (Ours)                & ResNet-50    & \bf 61.88   & \bf 67.02  & \bf 72.14   & \bf 74.25      \\ \hline
   
  \end{tabular}
  \end{center}
  \label{tab:voc_few}
\end{table}

The performance advantage of our approach in low-data regime is obvious. Our approach performs best among all the compared methods. It is superior to CCT \cite{CCT}, which 
also uses ResNet-50 as the backbone, and the improvement under all partition protocols are more than 10\%, from 12.15\% (using 732 labels) to 28.78\% (using 92 labels). Compared 
with other methods utilizing ResNet-101 as the backbone, CPCL still shows a significant performance improvement. Among the methods with ResNet-101, PseudoSeg \cite{pseudoseg} 
has the best overall performance, while CPCL achieves the increments of 1.84\%, 3.00\%, 1.52\%, and 4.28\% under different partition protocols. In addition, it can be seen 
that the mIoU (72.14\%) of our approach using 366 labeled images is comparable with that (73.23\%) of PseudoSeg using the full training set. The demand for labeled data can be 
effectively reduced.

\section{Conclusion and Discussion}

In this paper, we propose a novel semi-supervised learning approach for semantic segmentation, termed Conservative-Progressive Collaborative Learning (CPCL), to not only take the 
advantage of the high-quality labels but also make the full use of the large quantity of the unlabeled data. We realize the CPCL by the intersection and union pseudo supervision. 
The intersection pseudo supervision based on the agreement of the two paralleled networks is conservative and reliable, while the union pseudo supervision achieves the progressive 
exploration for the disagreement parts. They cooperate with each other and achieve the collaboration of the conservative evolution and the progressive exploration. To reduce the 
pseudo supervision noise, the confidence-based dynamic loss is proposed, which re-weights the loss on the basis of the prediction confidence. CPCL is simple, efficient and flexible. 
We have evaluated the proposed approach on two widely used datasets, Cityscapes and PASCAL VOC 2012. The extensive experimental results demonstrate the effectiveness of the 
CPCL, and show its state-of-the-art performance for semi-supervised semantic segmentation, especially in low-data regime. 

Although the performance is effectively improved, error predictions are unavoidable. It can be seen from the fourth row in Fig.~\ref{fig:cs_demo} that there are still segmentation errors 
between terrain and sidewalk. Union pseudo supervision is adopted for exploration, but the effect is limited when both the two networks are confidence with wrong predictions. 
In our opinion, the future works can be developed from the following perspectives.
Enhancing the quality of the pseudo supervision is the key to the good performance for the semi-supervised semantic segmentation. However, it is also a complicated and difficult 
task. We have tried several approaches, and most of them are mediocre. The main problem is how to estimate the reliability of the pseudo supervision without ground truth labels. 
In this paper, we leverage the agreement of the two paralleled models and the prediction confidence of the single model, but it is insufficient if model learned on labeled data 
is insufficient. Therefore, the evaluation of the pseudo supervision deserves further studied. Besides, the model coupling is another problem, which causes the peer-supervised 
networks to converge towards a wrong direction. Heterogeneous architecture is an effective way to reduce that. In this paper, we applied intersection pseudo supervision and union pseudo supervision 
to the two branches respectively. The status of the model coupling is usually empirically observed from the overlap ratio between the two models during training, 
which is usually small at the early training stage and increases during the later training. However, the overlap ratio may not be an effective indicator for model coupling, since 
we cannot distinguish the model coupling problem from the accurate models reaching a high degree of consensus with each other. Thus, the indicator for the model coupling problem is also worth 
to be further study.

\ifCLASSOPTIONcaptionsoff
  \newpage
\fi


\bibliographystyle{IEEEtran}
\bibliography{bib}

\begin{thebibliography}{10}
\providecommand{\url}[1]{#1}
\csname url@samestyle\endcsname
\providecommand{\newblock}{\relax}
\providecommand{\bibinfo}[2]{#2}
\providecommand{\BIBentrySTDinterwordspacing}{\spaceskip=0pt\relax}
\providecommand{\BIBentryALTinterwordstretchfactor}{4}
\providecommand{\BIBentryALTinterwordspacing}{\spaceskip=\fontdimen2\font plus
\BIBentryALTinterwordstretchfactor\fontdimen3\font minus
  \fontdimen4\font\relax}
\providecommand{\BIBforeignlanguage}[2]{{%
\expandafter\ifx\csname l@#1\endcsname\relax
\typeout{** WARNING: IEEEtran.bst: No hyphenation pattern has been}%
\typeout{** loaded for the language `#1'. Using the pattern for}%
\typeout{** the default language instead.}%
\else
\language=\csname l@#1\endcsname
\fi
#2}}
\providecommand{\BIBdecl}{\relax}
\BIBdecl

\bibitem{zhu2019parallel}
F.~Zhu, Y.~Lv, Y.~Chen, X.~Wang, G.~Xiong, and F.-Y. Wang, ``Parallel
  transportation systems: Toward iot-enabled smart urban traffic control and
  management,'' \emph{IEEE Trans. Intell. Transp. Syst.}, vol.~21, no.~10, pp.
  4063--4071, 2019.

\bibitem{Cityscapes}
M.~Cordts, M.~Omran, S.~Ramos, T.~Rehfeld, M.~Enzweiler, R.~Benenson,
  U.~Franke, S.~Roth, and B.~Schiele, ``The cityscapes dataset for semantic
  urban scene understanding,'' in \emph{Proc. IEEE Comput. Soc. Conf. Comput.
  Vision Pattern Recognit.}, 2016, pp. 3213--3223.

\bibitem{COCO}
T.-Y. Lin, M.~Maire, S.~Belongie, J.~Hays, P.~Perona, D.~Ramanan,
  P.~Doll{\'a}r, and C.~L. Zitnick, ``Microsoft coco: Common objects in
  context,'' in \emph{Lect. Notes Comput. Sci.}, 2014, pp. 740--755.

\bibitem{SSLEM}
Y.~Grandvalet and Y.~Bengio, ``Semi-supervised learning by entropy
  minimization,'' in \emph{Adv. neural inf. proces. syst.}, 2005, pp. 529--536.

\bibitem{ALS4}
W.-C. Hung, Y.-H. Tsai, Y.-T. Liou, Y.-Y. Lin, and M.-H. Yang, ``Adversarial
  learning for semi-supervised semantic segmentation,'' \emph{arXiv preprint
  arXiv:1802.07934}, 2018.

\bibitem{MT}
A.~Tarvainen and H.~Valpola, ``Mean teachers are better role models:
  Weight-averaged consistency targets improve semi-supervised deep learning
  results,'' in \emph{Adv. neural inf. proces. syst.}, 2017, p. 1196–1205.

\bibitem{VAT}
T.~Miyato, S.-I. Maeda, M.~Koyama, and S.~Ishii, ``Virtual adversarial
  training: A regularization method for supervised and semi-supervised
  learning,'' \emph{IEEE Trans Pattern Anal Mach Intell}, vol.~41, no.~8, pp.
  1979--1993, 2019.

\bibitem{9301261}
X.~Wang, D.~Kihara, J.~Luo, and G.-J. Qi, ``Enaet: A self-trained framework for
  semi-supervised and supervised learning with ensemble transformations,''
  \emph{IEEE Trans. on Image Process.}, vol.~30, pp. 1639--1647, 2021.

\bibitem{2019Semi}
G.~French, S.~Laine, T.~Aila, M.~Mackiewicz, and G.~Finlayson,
  ``Semi-supervised semantic segmentation needs strong, varied perturbations,''
  \emph{arXiv preprint arXiv:1906.01916}, 2019.

\bibitem{CCT}
Y.~Ouali, C.~Hudelot, and M.~Tami, ``Semi-supervised semantic segmentation with
  cross-consistency training,'' in \emph{Proc IEEE Comput Soc Conf Comput
  Vision Pattern Recognit}, 2020, pp. 12\,671--12\,681.

\bibitem{DualStudent}
Z.~Ke, D.~Wang, Q.~Yan, J.~Ren, and R.~Lau, ``Dual student: Breaking the limits
  of the teacher in semi-supervised learning,'' in \emph{Proc IEEE Int Conf
  Comput Vision}, 2019, pp. 6727--6735.

\bibitem{PLCB}
E.~Arazo, D.~Ortego, P.~Albert, N.~E. O'Connor, and K.~McGuinness,
  ``Pseudo-labeling and confirmation bias in deep semi-supervised learning,''
  in \emph{Proc Int Jt Conf Neural Networks}, 2020, pp. 1--1.

\bibitem{2021In}
M.~N. Rizve, K.~Duarte, Y.~S. Rawat, and M.~Shah, ``In defense of
  pseudo-labeling: An uncertainty-aware pseudo-label selection framework for
  semi-supervised learning,'' \emph{arXiv preprint arXiv:2101.06329}, 2021.

\bibitem{ke2019dual}
Z.~Ke, D.~Wang, Q.~Yan, J.~Ren, and R.~W. Lau, ``Dual student: Breaking the
  limits of the teacher in semi-supervised learning,'' in \emph{Proc. IEEE Int.
  Conf. Comput. Vision}, 2019, pp. 6727--6735.

\bibitem{qiao2018deep}
S.~Qiao, W.~Shen, Z.~Zhang, B.~Wang, and A.~Yuille, ``Deep co-training for
  semi-supervised image recognition,'' in \emph{Lect. Notes Comput. Sci.},
  2018, pp. 142--159.

\bibitem{2020DMT}
Z.~Feng, Q.~Zhou, Q.~Gu, X.~Tan, G.~Cheng, X.~Lu, j.~Shi, and L.~Ma, ``Dmt:
  Dynamic mutual training for semi-supervised learning,'' \emph{arXiv preprint
  arXiv:2004.08514}, 2020.

\bibitem{RML}
P.~Zhang, B.~Zhang, T.~Zhang, D.~Chen, and F.~Wen, ``Robust mutual learning for
  semi-supervised semantic segmentation,'' \emph{arXiv preprint
  arXiv:2106.00609}, 2021.

\bibitem{Pseudo-label}
D.-H. Lee, ``Pseudo-label: The simple and efficient semi-supervised learning
  method for deep neural networks,'' in \emph{Workshop on challenges in
  representation learning, ICML}, 2013, p. 896.

\bibitem{zou2018unsupervised}
Y.~Zou, Z.~Yu, B.~Kumar, and J.~Wang, ``Unsupervised domain adaptation for
  semantic segmentation via class-balanced self-training,'' in \emph{Lect.
  Notes Comput. Sci.}, 2018, pp. 297--313.

\bibitem{Naive-Student}
L.-C. Chen, R.~G. Lopes, B.~Cheng, M.~D. Collins, E.~D. Cubuk, B.~Zoph,
  H.~Adam, and J.~Shlens, ``Naive-student: Leveraging semi-supervised learning
  in video sequences for urban scene segmentation,'' in \emph{Lect. Notes
  Comput. Sci.}, 2020, pp. 695--714.

\bibitem{rethinking}
B.~Zoph, G.~Ghiasi, T.-Y. Lin, Y.~Cui, H.~Liu, E.~D. Cubuk, and Q.~V. Le,
  ``Rethinking pre-training and self-training,'' in \emph{Adv. neural inf.
  proces. syst.}, 2020, pp. 1--1.

\bibitem{ibrahim2020semi}
M.~S. Ibrahim, A.~Vahdat, M.~Ranjbar, and W.~G. Macready, ``Semi-supervised
  semantic image segmentation with self-correcting networks,'' in \emph{Proc.
  IEEE Comput. Soc. Conf. Comput. Vision Pattern Recognit.}, 2020, pp.
  12\,712--12\,722.

\bibitem{mendel2020semi}
R.~Mendel, L.~A. De~Souza, D.~Rauber, J.~P. Papa, and C.~Palm,
  ``Semi-supervised segmentation based on error-correcting supervision,'' in
  \emph{Lect. Notes Comput. Sci.}, 2020, pp. 141--157.

\bibitem{li2017triple}
C.~Li, T.~Xu, J.~Zhu, and B.~Zhang, ``Triple generative adversarial nets,'' in
  \emph{Adv. neural inf. proces. syst.}, 2017, pp. 4089--4099.

\bibitem{luo2018smooth}
Y.~Luo, J.~Zhu, M.~Li, Y.~Ren, and B.~Zhang, ``Smooth neighbors on teacher
  graphs for semi-supervised learning,'' in \emph{Proc. IEEE Comput. Soc. Conf.
  Comput. Vision Pattern Recognit.}, 2018, pp. 8896--8905.

\bibitem{springenberg2015unsupervised}
J.~T. Springenberg, ``Unsupervised and semi-supervised learning with
  categorical generative adversarial networks,'' \emph{arXiv preprint
  arXiv:1511.06390}, 2015.

\bibitem{souly2017semi}
N.~Souly, C.~Spampinato, and M.~Shah, ``Semi supervised semantic segmentation
  using generative adversarial network,'' in \emph{Proc. IEEE Int. Conf.
  Comput. Vision}, 2017, pp. 5689--5697.

\bibitem{8935407}
S.~Mittal, M.~Tatarchenko, and T.~Brox, ``Semi-supervised semantic segmentation
  with high- and low-level consistency,'' \emph{IEEE Transactions on Pattern
  Analysis and Machine Intelligence}, vol.~43, no.~4, pp. 1369--1379, 2021.

\bibitem{athiwaratkun2018there}
B.~Athiwaratkun, M.~Finzi, P.~Izmailov, and A.~G. Wilson, ``There are many
  consistent explanations of unlabeled data: Why you should average,'' in
  \emph{Int. Conf. Learn. Represent., ICLR}, 2018, pp. 1--1.

\bibitem{laine2016temporal}
S.~Laine and T.~Aila, ``Temporal ensembling for semi-supervised learning,'' in
  \emph{Int. Conf. Learn. Represent., ICLR - Conf. Track Proc.}, 2017, pp.
  1--1.

\bibitem{kim2020structured}
J.~Kim, J.~Jang, and H.~Park, ``Structured consistency loss for semi-supervised
  semantic segmentation,'' \emph{arXiv preprint arXiv:2001.04647}, 2020.

\bibitem{MixMatch}
D.~Berthelot, N.~Carlini, I.~Goodfellow, A.~Oliver, N.~Papernot, and C.~Raffel,
  ``Mixmatch: A holistic approach to semi-supervised learning,'' in \emph{Adv.
  neural inf. proces. syst.}, 2019, p. 5049–5059.

\bibitem{berthelot2019remixmatch}
D.~Berthelot, N.~Carlini, E.~D. Cubuk, A.~Kurakin, K.~Sohn, H.~Zhang, and
  C.~Raffel, ``Remixmatch: Semi-supervised learning with distribution alignment
  and augmentation anchoring,'' \emph{arXiv preprint arXiv:1911.09785}, 2019.

\bibitem{devries2017improved}
T.~DeVries and G.~W. Taylor, ``Improved regularization of convolutional neural
  networks with cutout,'' \emph{arXiv preprint arXiv:1708.04552}, 2017.

\bibitem{zhang2017mixup}
H.~Zhang, M.~Cisse, Y.~N. Dauphin, and D.~Lopez-Paz, ``mixup: Beyond empirical
  risk minimization,'' \emph{arXiv preprint arXiv:1710.09412}, 2017.

\bibitem{CPS}
X.~Chen, Y.~Yuan, G.~Zeng, and J.~Wang, ``Semi-supervised semantic segmentation
  with cross pseudo supervision,'' in \emph{Proc. IEEE Comput. Soc. Conf.
  Comput. Vision Pattern Recognit.}, 2021, pp. 2613--2622.

\bibitem{sohn2020fixmatch}
K.~Sohn, D.~Berthelot, C.-L. Li, Z.~Zhang, N.~Carlini, E.~D. Cubuk, A.~Kurakin,
  H.~Zhang, and C.~Raffel, ``Fixmatch: Simplifying semi-supervised learning
  with consistency and confidence,'' \emph{arXiv preprint arXiv:2001.07685},
  2020.

\bibitem{pseudoseg}
Y.~Zou, Z.~Zhang, H.~Zhang, C.-L. Li, X.~Bian, J.-B. Huang, and T.~Pfister,
  ``Pseudoseg: Designing pseudo labels for semantic segmentation,'' \emph{arXiv
  preprint arXiv:2010.09713}, 2020.

\bibitem{park2018adversarial}
S.~Park, J.~Park, S.-J. Shin, and I.-C. Moon, ``Adversarial dropout for
  supervised and semi-supervised learning,'' in \emph{AAAI Conf. Artif.
  Intell., AAAI}, 2018, pp. 3917--3924.

\bibitem{GCT}
Z.~Ke, D.~Qiu, K.~Li, Q.~Yan, and R.~W. Lau, ``Guided collaborative training
  for pixel-wise semi-supervised learning,'' in \emph{Lect. Notes Comput.
  Sci.}, 2020, pp. 429--445.

\bibitem{sun2020mining}
G.~Sun, W.~Wang, J.~Dai, and L.~Van~Gool, ``Mining cross-image semantics for
  weakly supervised semantic segmentation,'' in \emph{Lect. Notes Comput.
  Sci.}, 2020, pp. 347--365.

\bibitem{5504216}
M.~Makitalo and A.~Foi, ``Optimal inversion of the anscombe transformation in
  low-count poisson image denoising,'' \emph{IEEE Trans. on Image Process.},
  vol.~20, no.~1, pp. 99--109, 2011.

\bibitem{7577781}
R.~Xiong, H.~Liu, X.~Zhang, J.~Zhang, S.~Ma, F.~Wu, and W.~Gao, ``Image
  denoising via bandwise adaptive modeling and regularization exploiting
  nonlocal similarity,'' \emph{IEEE Trans. on Image Process.}, vol.~25, no.~12,
  pp. 5793--5805, 2016.

\bibitem{pascalvoc}
M.~Everingham, S.~A. Eslami, L.~Van~Gool, C.~K. Williams, J.~Winn, and
  A.~Zisserman, ``The pascal visual object classes challenge: A
  retrospective,'' \emph{Int. J. Comput. Vision}, vol. 111, no.~1, pp. 98--136,
  2015.

\bibitem{SBD}
B.~Hariharan, P.~Arbeláez, L.~Bourdev, S.~Maji, and J.~Malik, ``Semantic
  contours from inverse detectors,'' in \emph{Proc. IEEE Int. Conf. Comput.
  Vision}, 2011, pp. 991--998.

\end{thebibliography}

\end{document}